%% file: main.tex
\documentclass[12pt]{article} 
\usepackage{amsmath,amsthm,amssymb,bm,mathrsfs}
\usepackage{caption,enumerate,listings,setspace}
\usepackage[colorlinks,linkcolor=red,anchorcolor=blue,citecolor=blue]{hyperref}
\usepackage[margin=1in]{geometry}
\usepackage[title]{appendix}
\usepackage{enumitem}
\theoremstyle{plain}
\usepackage{booktabs}
\usepackage[numbers]{natbib}
\usepackage{array}
\usepackage{algorithm}
\usepackage{algpseudocode}
\usepackage{makecell}

\input{def-smile}

\input{front}

\title{\textbf{Advancing Retail Data Science: Comprehensive Evaluation of Synthetic Data}} 

\author{
Yu Xia \thanks{Machine Learning Engineer, Bain \& Company, CA, 94158. Email: yu.xia@bain.com}, 
Chi-Hua Wang \thanks{Postdoctoral Scholar, Department of Statistics \& Data Science,  UCLA, CA, 90095. Email: chihuawang@ucla.edu},
Joshua Mabry \thanks{Sr. Director of Data Science \& ML Engineering, Bain \& Company, CA, 94158. Email: joshua.mabry@bain.com}, 
Guang Cheng \thanks{Professor, Department of Statistics and Data Science, UCLA, CA, 90095. Email: guangcheng@ucla.edu}
}

\begin{document}

\maketitle

\begin{abstract}
The evaluation of synthetic data generation is crucial, especially in the retail sector where data accuracy is paramount. This paper introduces a comprehensive framework for assessing synthetic retail data, focusing on fidelity, utility, and privacy. Our approach differentiates between continuous and discrete data attributes, providing precise evaluation criteria.
Fidelity is measured through stability and generalizability. Stability ensures synthetic data accurately replicates known data distributions, while generalizability confirms its robustness in novel scenarios. Utility is demonstrated through the synthetic data's effectiveness in critical retail tasks such as demand forecasting and dynamic pricing, proving its value in predictive analytics and strategic planning. Privacy is safeguarded using Differential Privacy, ensuring synthetic data maintains a perfect balance between resembling training and holdout datasets without compromising security.
Our findings validate that this framework provides reliable and scalable evaluation for synthetic retail data. It ensures high fidelity, utility, and privacy, making it an essential tool for advancing retail data science. This framework meets the evolving needs of the retail industry with precision and confidence, paving the way for future advancements in synthetic data methodologies. 
\end{abstract}


\clearpage
\section{Introduction}
\label{sec:intro}

In the rapidly evolving field of data science, the evaluation of synthetic data generation frameworks has become paramount, especially within the retail sector. This paper introduces a comprehensive framework for assessing synthetic retail data, focusing on three critical dimensions: fidelity, utility, and privacy. Our framework distinguishes between continuous and discrete attributes within retail datasets, providing clear methodologies for their evaluation.

Firstly, fidelity is evaluated through stability and generalizability \cite{tao2024discriminative}. Stability measures how well synthetic retail data replicates known data distributions, highlighting the robustness of models in familiar scenarios. Generalizability, on the other hand, assesses the performance of synthetic data in novel contexts, ensuring that the generated data can effectively extend beyond its training parameters. This is particularly important in retail, where market trends and consumer behavior can shift rapidly.

Secondly, the utility of synthetic retail data is scrutinized by its applicability to real-world tasks. In the retail sector, accurate demand forecasting and dynamic pricing \cite{wang2023online, liu2022non} are pivotal for operational efficiency and profitability. Our evaluation framework demonstrates how synthetic datasets can effectively support these core functions, making them indispensable for predictive analytics and strategic decision-making in retail.

Finally, privacy is assessed using Differential Privacy and related metrics. We compare the proximity of synthetic datasets to both training and holdout datasets to ensure balanced privacy guarantees. A well-balanced synthetic dataset should approximate both datasets equally, indicating robust privacy protection without compromising data utility. This aspect is critical in retail, where customer data privacy is a significant concern.

We apply our framework to evaluate generative AI models trained with the Complete Journey dataset \cite{dunnhumby_complete_2014}. Our results affirm that our evaluation framework provides a robust pipeline for large-scale assessments of synthetic retail data generation models. This framework not only ensures high fidelity and utility but also maintains stringent privacy standards. Consequently, it offers a solid foundation for future improvements in synthetic data generation and evaluation methodologies within the retail sector.

This paper concludes that with our framework, synthetic retail data can be reliably utilized for various applications, offering a scalable solution for the ever-growing demands of data privacy and utility in retail data science.

\subsection{Background and Motivation}

In the retail industry, the challenges of data privacy and availability are significant obstacles. Synthetic data, which is artificially generated rather than obtained from real-world events, provides a compelling solution to these issues. One primary challenge in retail is protecting customer privacy while leveraging data for analysis and decision-making. Synthetic data addresses this by mimicking real data without exposing sensitive customer information, and maintaining statistical properties and patterns found in actual data. This allows retailers to perform robust analyses and model training without risking data breaches or violating privacy regulations.

Moreover, obtaining large volumes of high-quality data can be difficult, especially when dealing with new products or services where historical data is sparse or non-existent. Public datasets are notably smaller than standard industry datasets. They are generally collected under biased and often undisclosed marketing policies, and they lack many critical fields needed for accurate customer behavior modeling\cite{de_biasio_systematic_2023}. Moreover, business constraints and fairness concerns restrict the potential for aggressive experimentation across the marketing mix. Synthetic data generation overcomes these by creating abundant and varied datasets that reflect potential future scenarios or underrepresented cases. This capability is crucial for training machine learning models, which require large datasets to perform effectively. Additionally, synthetic data can help mitigate biases present in real data, leading to more fair and accurate models.

\subsection{Objectives and Contributions}

Developing a robust evaluation framework for synthetic data in retail is essential to ensure the validity and utility of the data. Without rigorous evaluation, synthetic data may fail to accurately reflect the complexities of real-world scenarios, leading to misleading insights and poor decision-making. A strong evaluation framework involves several critical components: assessing the statistical similarity between synthetic and real data, evaluating the impact on model performance, and ensuring that synthetic data preserves essential patterns and relationships. Thus, we propose a standardized evaluation framework for retail synthetic datasets from three aspects: fidelity, utility, and privacy.

\begin{figure}[H]
    \centering
    \includegraphics[width=0.85\linewidth]{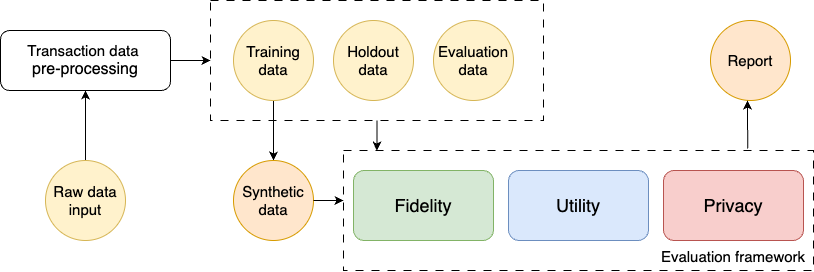}
    \caption{The framework diagram of our synthetic retail data evaluation pipeline. Section \ref{subsec:data-split} explains the purpose and method to split transaction data. Section \ref{subsec:define-fidelity} defines detailed metrics for fidelity assessment, i.e. Wasserstein distance,  Pearson correlation, etc. Section \ref{subsec:define-utility} defines the tasks for utility assessment, i.e. classification accuracy, product association, etc. Section\ref{subsec:define-privacy} explains the metrics for privacy assessment, i.e. distance to the closest record.}
    \label{fig:framework}
\end{figure}

Such a framework (Figure \ref{fig:framework}) ensures that synthetic data is not only statistically similar to real data but also useful for practical applications in the retail sector. This process helps identify any discrepancies and areas where synthetic data may fall short, guiding improvements in data generation methods. Ultimately, a robust evaluation framework builds trust in synthetic data, making it a reliable resource for retailers. In this way, we ensure a safe and scalable way to generate high-quality synthetic data while maintaining privacy compliance.

% %\clearpage
\section{Related Work}
\label{sec:related_works}

\subsection{Existing Evaluation Frameworks}

Several general frameworks have been proposed to gauge the efficacy of synthetic data previously. A sample-level metric framework evaluates generative models through fidelity and utility lenses, facilitating the identification of discrepancies and similarities between real and synthetic datasets \cite{alaa2022faithful}. Another methodology emphasizes auditing and generating synthetic data with controllable trust trade-offs, allowing customization based on specific requirements \cite{belgodere2023auditing}. Further exploration of synthetic data generation discusses its benefits and limitations across various contexts, particularly in creating a practical approach for deployment \cite{jordon_synthetic_2022}.

Fidelity assessment ensures synthetic data retains the essential characteristics and patterns of real data. Previous work proposed a holdout-based empirical assessment method for mixed-type synthetic data, highlighting the importance of maintaining the statistical properties and variability inherent in the original dataset \cite{platzer2021holdout}. On the metric aspect, Sajjadi et al introduced a definition of precision and recall for distribution and quantified distribution similarity not just with one-dimensional score, like total variation \cite{sajjadi2018assessing}.

Utility evaluation focuses on the synthetic data's performance in downstream tasks. Xu et al. established a basis of relevant utility theory in a statistical learning framework and introduced metrics of generalization and ranking of models trained on synthetic data \cite{xu2023utility}. It considers two utility metrics: generalization and ranking of models trained on synthetic data.  There was also empirical work, for example, emphasizing generative model selection based on performance in fraud detection\cite{cheng2024downstream}. Additionally, Hsieh et al. (2024) adopted a data-centric perspective to improve both the fidelity and utility of synthetic credit card transaction time series \cite{hsieh2024improve}. Liu et al. explore utility in dynamic pricing models, demonstrating how synthetic data can support robust pricing strategies in fluctuating market conditions \cite{liu2022utility}.

Privacy is a central concern in synthetic data generation. A formal framework for detecting data-copying in generative models ensures synthetic data does not replicate real data points \cite{bhattacharjee2023data}, where the author also provides the requirement of minimum sample size for reliable detection. Meehan's work proposed a three-sample test to solve the same issue of data-copying in generative models \cite{meehan2020three}. BadGD addresses the vulnerabilities of gradient descent algorithms through strategic backdoor attacks to safeguard data privacy \cite{wang2024badgd}. Furthermore, Chen et al. systematically summarize all approaches for differentially private data publishing to conduct reproducible downstream analysis while preserving data privacy \cite{chen2023unified}. Tools like TAPAS provide adversarial privacy auditing. A review of privacy measurement practices for tabular synthetic data includes a comprehensive list of  privacy metrics \cite{boudewijn2023privacy}, such as Differential Privacy,  k-Anonymity,  Plausible Deniability, etc.

However, to our knowledge, there is no empirical work conducting a full set of fidelity, utility, and privacy assessments on generative AI models with retail transaction data.

\subsection{Synthetic Data in Retail}
Synthetic data can transform the retail industry by enhancing various operational and analytical processes while ensuring customer privacy. It enables comprehensive customer analytics and segmentation without compromising personal data, aiding in the development of targeted marketing strategies. In supply chain optimization, synthetic data simulates different scenarios to help forecast demand, optimize inventory, and improve logistics. Product recommendation systems can benefit from data augmentation with synthetic datasets for extensive training, ensuring accurate and relevant recommendations that enhance customer experience. Furthermore, synthetic data allows safe data sharing with third parties and partners under privacy regulations, facilitating collaborative projects and compliance checks without using real data, thus fostering innovation while safeguarding privacy.

Data practitioners have widely recognized the value of synthetic data for accelerating the development of AI systems and started to emphasize building generative models of the data in its raw tabular forms, instead of modeling features derived from transformed data \cite{patki_synthetic_2016, jordon_synthetic_2022}. Researchers identify two different paths for synthetic data generation. One is the black-box style of privacy-preserving modeling techniques (such as Generative Adversarial Networks, Variational Autoencoders, and Bayesian Networks) \cite{wilde_foundations_2021}. For example, Athey generated synthetic data for the evaluation of causal effects estimators with Wasserstein Generative Adversarial Networks \cite{athey_estimating_2020}. These privacy-preserving modeling techniques are powerful when sufficient historical data is available to learn an accurate data-generating process. However, when public retail datasets are rather scarce, the other style that needs domain knowledge in retail shines. Statisticians simplify the complexity of real-world data and specify structural causal models to generate synthetic data. For example, prior work simulated either category choice or the full life-cycle of customer shopping decisions based on a nested logit model \cite{andrews_comparison_2011, xia_retailsynth_2023}.

However, no matter which technique is used for synthetic data generation, there is currently no well-defined evaluation framework specifically designed to assess synthetic retail data, highlighting a crucial gap in ensuring data fidelity, utility, and privacy. The retail industry particularly requires effective data to conduct analysis, such as price optimization\cite{chen2022statistical, Chen_2020}, basket analysis \cite{van2023next, Ruiz_2020}, customer lifetime value \cite{romero_partially_2013, ekinci_using_2014}, and demand forecasting \cite{wan_modeling_2017, gabel2022product}. Given the complexity and importance of these tasks, careful evaluation of synthetic data is essential to preserve the quality of insights derived from these analyses, ensuring that strategic decisions are based on accurate and reliable information.

\subsection{Complete Journey Dataset}
\label{subsec:complete_journey_dataset}

The Dunnhumby Complete Journey Dataset \cite{dunnhumby_complete_2014} represents a comprehensive and meticulously curated collection of retail transaction data, providing deep insights into consumer purchase behaviors and patterns. Compiled from a vast array of shopping experiences, this dataset encompasses detailed records of customer interactions, including basket-level transaction details, promotional influences, and loyalty program participation. We utilize three main tables in our paper from the dataset: (1) transaction table, (2) customer demographics table, and (3) product hierarchy table. We merge these three tables with schema explained in Table \ref{tab:raw-data-input-schema} to build the raw data input to the proposed framework in Figure \ref{fig:framework}, and cover the pre-processing details in section \ref{subsec:dataset-eda}.

\begin{table}[]
    \centering
    \begin{tabular} [width=1.0\linewidth]{cccc}
    \toprule
         Variable&  Type &Description &Source\\
    \midrule
         household\_id& integer& Customer unique identifier& 1 \& 2\\
         product\_id& integer& Product unique identifier& 1 \& 3\\
         day& integer& Transaction day& 1\\
         week& integer& Transaction week & 1\\
         quantity& integer& Purchased units& 1\\
         sales\_value& float& Revenue& 1\\
         ...& ...& ...&\\
         \midrule
         age& string& Customer age group&2\\
         household\_size& string& Number of family members&2\\
         ...& ...& ...&\\
        \midrule
         department& string& Product department& 3\\
         brand& string& "national" or "private"&3\\
         ...& ...& ...&\\
         \bottomrule
         \end{tabular}
    \caption{Partial schema of merged Complete Journey dataset as an example, where (1) in the source column represent the transactions table, (2) for the customer demographics table, and (3) for the product hierarchy table. See Sec. \ref{subsec:complete_journey_dataset} for full details.}
    \label{tab:raw-data-input-schema}
\end{table}

\section{The Evaluation Framework}

In this session, we propose an empirical assessment framework to evaluate generative AI models for retail synthetic data generation. Our evaluation method is distinguished by its tripartite composition of synthetic data \textbf{Fidelity}, \textbf{Utility}, and \textbf{Privacy} metrics. The defining characteristic of this method is its adaptability and model-free nature, allowing it to be deployed independently of domain-specific knowledge or preconceived notions. Utilizing non-parametric measures, our data-centric evaluation provides a systematic review of an array of black-box synthetic data solutions, examining whether generated data is practical, safe, and broadly applicable. The objective underlining this methodology is to build transparency, enhance confidence in data generators, and further incentivize industries to leverage synthetic data for innovation in the modern data-driven world.

\subsection{Train-Holdout-Eval Split}\label{subsec:data-split}
To robustly evaluate the generalizability of a data synthesizer, we employ a random split of the available records into three distinct datasets: a training dataset $T$, a holdout dataset $H$ \cite{platzer2021holdout}, and an additional evaluation dataset $E$, where the evaluation dataset $E$ is only used for assessing model utility as a evaluation set (Figure \ref{fig:framework}). The training dataset $T$ is exclusively used to train the synthesizer, while the holdout dataset $H$ remains untouched during the synthetic data generation process. By exposing only the training dataset $T$ to the synthesizer, we generate a synthetic dataset $S$ of the same size as $T$. The isolated holdout dataset $H$ can then serve as a benchmark to assess the synthesizer’s ability to generalize beyond the data it was trained on.

To demonstrate the model generalizability, we compare the metrics obtained from both the holdout dataset $H$ and the synthetic dataset $S$. If the holdout dataset $H$ attains better metrics than the synthetic dataset $S$, the synthesizer has missed some underlying patterns presented in the data. Conversely, suppose the synthetic dataset S achieves superior metrics. In that case, this indicates potential overfitting by the synthesizer to the training dataset $T$. Ideally, we aim for the metrics from both datasets to be as close as possible, reflecting balanced generalization and reliable synthetic data generation.

\subsection{Measuring Synthetic Data Fidelity} \label{subsec:define-fidelity}

We treat holdout and synthetic datasets as separate data sources to evaluate fidelity metrics against the training dataset. The design of fidelity measurement is motivated by visualizing joint distributions and marginal distributions to discover patterns. 

\textbf{Similarity of Marginal Distribution}  One critical part of exploratory data analysis is to demonstrate the distribution of numerical features. This involves plotting histograms, density plots, and cumulative distribution functions to visualize how numerical data is spread across different ranges. A robust generative synthesizer must accurately learn and replicate these numerical distributions, ensuring that the synthetic data mirrors the real-world data in terms of central tendencies, variability, and distribution shape. Furthermore, we can also derive additional features from primitive columns and test their distribution similarities. This includes calculating ratios, differences, and other mathematical transformations to extract business insights. By assessing the distribution of these derived features, we can further evaluate the synthesizer's stability and robustness, ensuring that it captures intricate relationships and patterns within the data. For both primitive numerical features and derived numerical features, we report Wasserstein distance \cite{ruschendorf1985wasserstein} to measure distribution similarities, where the small value indicates the synthetic dataset is closely attached to the real dataset.

Another important aspect is to check the distribution of categorical features. This involves visualizing the frequency of each category, cross-tabulations, and bar plots to understand the distribution of categorical data. A competent generative synthesizer must also learn these categorical distributions accurately. It should preserve the proportions and relationships among categories, ensuring that the synthetic data accurately represents the categorical structures observed in the real dataset. To quantify the degree of similarity, we compute the Jensen-Shannon distance \cite{briet2009properties} and expect a small value as an indicator of an excellent synthesizer.

\textbf{Similarity of Joint Distribution} Besides capturing the distribution of a single attribute, a synthesizer with high fidelity should also be able to identify multivariate combinations and relationships among the set of attributes, assessing how pairs of features interact and co-vary. We compute the Pearson correlation matrix for number-to-number interaction, Theil's U matrix for category-to-category interaction, and the correlation ratio matrix for number-to-category interactions. To verify if the synthesizer understands feature interactions and dependencies, we compute the L2 distance of flattened correlation arrays between the training dataset and the synthetic dataset or the holdout dataset. This step is vital for applications where the relationship between variables significantly impacts outcomes, such as customer segmentation and market basket analysis in the retail industry. Ensuring that these joint distributions are faithfully replicated in the synthetic data guarantees that the model maintains the integrity of multivariate relationships, providing a more comprehensive and realistic representation of the underlying data structure.

\subsection{Measuring Synthetic Data 
 Utility}\label{subsec:define-utility}
In evaluating the efficacy of generative models for retail synthetic data generation, a critical consideration is the preservation of Machine Learning (ML) utility. This step entails formulating a classification task $f: X \rightarrow Y$ using a predefined dataset, enabling a comprehensive assessment of how well-synthesized data can replicate real-world data's utility in predictive modeling. The evaluation framework is meticulously designed to ensure a robust comparison of model performance on both utility and generalizability.

To achieve this, we train machine learning models separately using the training dataset $T$, holdout dataset $H$, and synthetic dataset $S$. This approach allows us to systematically assess the performance of each model on the same evaluation set, referred to as evaluation dataset $E$. The model trained with $T$, $f_T$, provides a baseline for understanding performance metrics under standard conditions. Conversely, the model trained with $H$, $f_H$, offers insights into the model's behavior when exposed to unseen real-world data, thereby indicating its generalizability. Furthermore, the model trained with $S$, $f_S$, in this evaluative procedure enables a direct comparison of how well the generative models could replicate actual data characteristics and maintain predictive accuracy.

By testing $f_T, f_H, f_S$ on evaluation dataset $E$ and computing metrics like accuracy, F1, ROC, precision, and recall, we are able to systematically quantify and compare the utility of data generated by various generative AI models. This empirical assessment framework not only facilitates a granular understanding of each model's performance but also highlights the strengths and limitations of generative AI in capturing complex data patterns crucial for prediction tasks in the retail sector.

\subsection{Measuring Synthetic Data Privacy}\label{subsec:define-privacy}

Privacy is a paramount concern in the realm of synthetic tabular data generation, primarily due to the sensitive nature of information often contained within retail datasets. The generation of synthetic data aims to mitigate the risk of disclosing private or proprietary information while still enabling valuable data-driven insights. To rigorously evaluate the privacy-preserving capabilities of generative AI models, we compute the Distance to Closest Record (DCR) , with L1 distance as the definition of distance between two records. Specifically, we assess the DCR from the synthetic data $S$ to the training data $T$, and from the holdout data $H$ to the training data $T$. The DCR quantifies the likelihood of synthetic data points being too similar to actual data points, thereby posing a privacy threat. A high DCR value indicates effective anonymization.

Additionally, we introduce a metric termed the \textbf{Closest Cluster Ratio (CCR)} further to scrutinize the privacy and generalizability of synthetic data. The CCR measures the proportion of synthetic data points that are closer to the training dataset compared to the holdout dataset, ranging from $[0, 1]$. Ideally, the values of CCR should be as low as possible, indicating that synthetic data points are not a close copy of the training dataset. A CCR close to 1 signals an overfitting generative model, highlighting the necessity for continuous refinement in synthetic data generation techniques. 

By combining DCR and CCR metrics, we can provide deep insights into how effectively synthetic data can protect sensitive information, thereby fostering trust and reliability in the deployment of generative AI solutions in real-world retail scenarios.

\section{Evaluation Results}

To demonstrate the proposed framework (Figure \ref{fig:framework}), we conducted an empirical assessment on the open-source Complete Journey dataset \cite{dunnhumby_complete_2014} of retail transactions from frequent customers in a retail grocery store (accessed from \textit{completejourney-py} \footnote{https://pypi.org/project/completejourney-py/}). The dataset documents the purchasing patterns of more than two thousand households over a one-year period, who frequently shop at the retailer.  

We examined 5 generative models to produce the synthetic datasets: GAN-based tabular generative models ((1) CTGAN\cite{ctgan}, (2) AutoGAN) and Diffusion-based tabular generative models ((3) TabDDPM\cite{kotelnikov2023tabddpm}, (4) StasyAutoDiff\cite{suh2023autodiff}, (5) TabAutoDiff\cite{suh2023autodiff}). Specifically, we implemented AutoGAN by preparing input features with AutoDiff \cite{suh2023autodiff} and training a GAN\cite{goodfellow2020generative} using Torch \cite{Ansel_PyTorch_2_Faster_2024}. Unless otherwise specified, models are cloned from the cited repositories and the training features are prepared according to encoding methods stated in the corresponding paper.

\subsection{Data Description and Analysis} \label{subsec:dataset-eda}
\paragraph{\textbf{Data Preprocess}} The raw transaction data includes approximately 1.47 million transactions and a wide range of about 92,000 products. The dataset presents a detailed category hierarchy that includes product department, product category, and product type. It also offers comprehensive customer demographics, such as age, income, household size, and marital status. Key transaction information, like item quantity, transaction sales amount, and discounts, are documented, enabling the calculation of unit prices and discounts \ref{tab:raw-data-input-schema}. We followed the same pre-process procedure shown in RetailSynth \cite{xia_retailsynth_2023} to remove seasonality effects, by cleaning out unregistered customers, excluding transactions with non-positive transactions, de-duplicating the product catalog, removing infrequent products, and aggregating weekly transactions for each customer. This is a typical procedure for optimizing marketing spend, customer lifetime value calculation, etc. To further increase the effective data points for each customer, we clustered customers by their demographic information and ended up with a weekly retail transaction with about 251,000 records from 6000 products and 400 customer clusters.

\begin{figure}[h]
    \centering
    \includegraphics[width=0.7\linewidth]{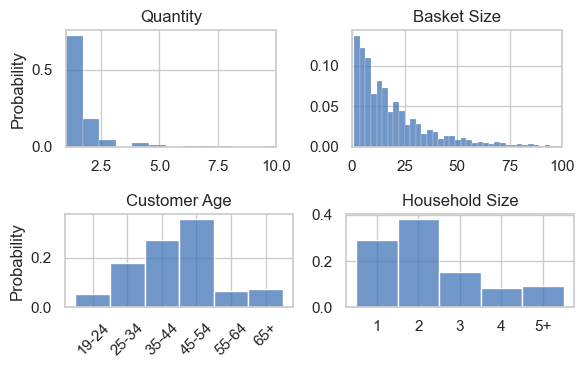}
    \caption{Selected univariate distributions for dataset “Complete Journey” illustrate diverse distributional patterns encountered in real-world datasets (see section \ref{paragraph:marginal-dist}).}
    \label{fig:eda}
\end{figure}

\paragraph{\textbf{Data Analysis}} To generate more customer- and product-level insights, we calculated derived features from the processed dataset, such as product purchase probability, store visit probability, basket size, etc. Figure \ref{fig:eda} exhibits two numeric columns on the top row, showing the skewed distributions of native feature, quantity, and derived feature, basket size, in the real-world retail transaction dataset. The distribution of quantities purchased tends to be positively skewed because most customers typically buy products in small quantities. Bulk purchases are less frequent, leading to a long tail on the right side of the distribution. The "Basket Size" subplot shows the probability distribution of the total number of items in a customer's basket. The distribution is right-skewed, indicating that while most transactions have a lower total basket size, a few transactions involve significantly higher total purchases. This is typical in retail, where a small number of premium customers can drive a substantial portion of revenue. \label{paragraph:marginal-dist}

% \begin{figure}%
%     \centering
%     \subfloat[\centering Selected univariate distributions]{{\includegraphics[width=0.48\textwidth]{assets/eda.png} }}
%     \qquad
%     \subfloat[\centering Marginal distribution of basket size ]{{\includegraphics[width=0.48\textwidth]{assets/basket_size_by_household_sze.png} }}
%     \caption{(a) shows selected univariate distributions for dataset “Complete Journey” illustrate diverse distributional patterns encountered in real-world datasets (see section \ref{paragraph:marginal-dist}). (b) shows the marginal distribution of basket size by different household sizes. Customers with more family members in the household tend to buy more products in one visit (see section \ref{paragraph:joint-dist}).}%
%     \label{fig:example}%
% \end{figure}

Similarly, the bottom row details categorical distributions for "Customer Age", and "Household Size". Our dataset has a slight concentration of customers in the middle age groups (e.g., 35-44 and 45-54), suggesting that middle-aged consumers form a large portion of the customer base. However, younger (19-24, 25-34) age groups are also well-represented. Household size describes the number of members per household. The distribution peaks at household sizes of 2 and 3, indicating that most customers come from small to medium-sized households.

\begin{figure}[h]
    \centering    \includegraphics[width=0.7\linewidth]{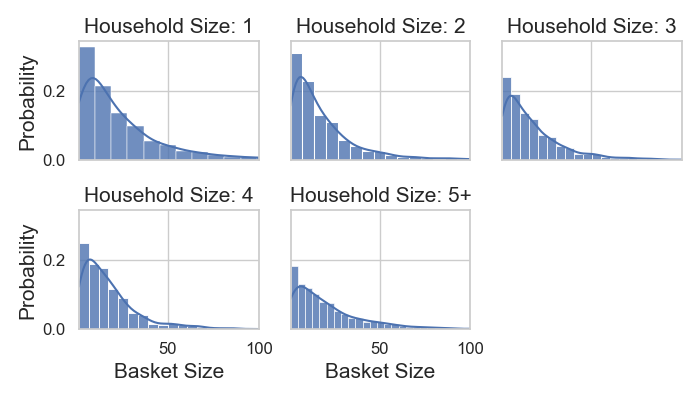}
    \caption{Marginal distribution of basket size by different household sizes. Customers with more family members in the household tend to buy more products in one visit (see section \ref{paragraph:joint-dist}).}
    \label{fig:dist_by_group}
\end{figure}

We also looked into multivariate combinations to explore relationships among the pair of columns. For example, Figure \ref{fig:dist_by_group} demonstrates that as household size increases, the distribution of basket sizes progressively shifts to the right. In single-member households, purchases predominantly consist of smaller basket sizes. As household size grows from two to three and onward to five plus members, the distribution begins to show a higher density of larger basket sizes. This shift to the right indicates that larger households tend to buy more in a single transaction, reflecting their greater consumption needs. \label{paragraph:joint-dist}

\begin{figure}[h]
    \centering
    \includegraphics[width=0.85\linewidth]{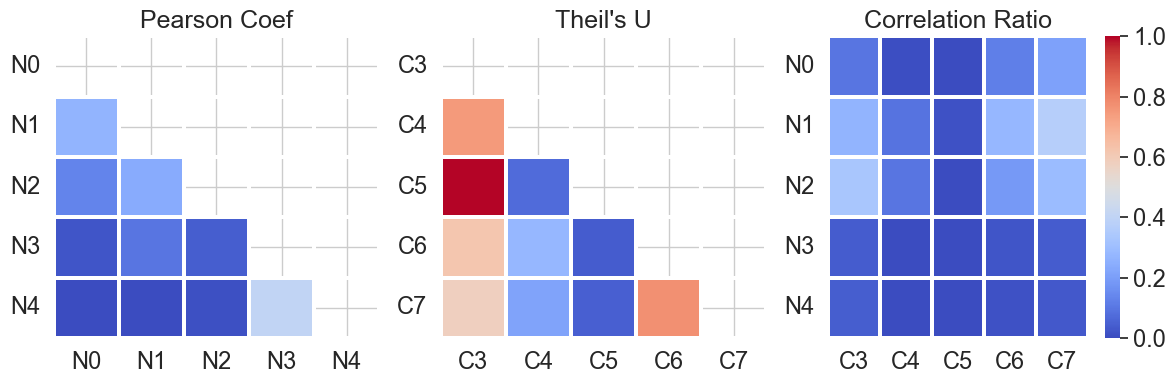}
    \caption{Correlation of selected numeric and categorical distributions for dataset “Complete Journey” illustrating contextual relationships observed in the real-world dataset (see Sec. \ref{paragraph:joint-dist-corr}).}
    \label{fig:corr-eda}
\end{figure}

Figure \ref{fig:corr-eda} shows a more comprehensive view on the Pearson correlation coefficient for numerical-numerical feature relationships, Theil’s U statistic for assessing dependence between categorical-categorical features, and the correlation ratio for categorical-numerical feature associations. For example, a high positive correlation between manufacturer id (C3) and department (C4), see appendix \ref{appendix:column-mapping} for a full list column mapping. Department, manufacturer, product category, product type, and package size are strongly associated with indicating a given product. \label{paragraph:joint-dist-corr}

\subsection{Synthetic Data Fidelity Assessment}
\label{subsec:fidelity_assessment}

\paragraph{\textbf{Distributional similarity overview}}  For the holdout dataset and synthetic datasets, we computed the average Wasserstein distance for numerical columns, the average Jensen-Shannon distance for categorical columns, and Euclidean distances of the Pearson correlation matrix, Theil's U matrix, and correlation ratio matrix against the training dataset. Table \ref{tab:fidelity-table} presents a comprehensive evaluation of various generative AI models for retail synthetic data generation, highlighting the diverse strengths and weaknesses of each model. 

\begin{table}[!htb]
    \centering
    \begin{tabular}[width=1.0\linewidth]{clllll}
\toprule
 & \multicolumn{2}{c}{Marginal}&\multicolumn{3}{c}{Joint}\\
 \cmidrule(lr){2-4} \cmidrule(lr){5-6}
 & Num&   Cat&Num-Num&Cat-Cat&Num-Cat\\
 \midrule
 Holdout & 0.04& 0.38& 0.45& 0.04& 0.04\\
 \midrule
 CTGAN& 2.24& 0.46& \textbf{0.49}& 4.06& \textbf{0.75}\\
 AutoGAN& 1646.88&   0.41&2.13&8.15&4.28\\
 TabDDPM& 5.36&  \textbf{0.38}& 0.85& 3.79& 1.22\\
 StasyAutoDiff& 7.55&\textbf{0.38}&1.18& 3.89& 1.59\\
 TabAutoDiff& \textbf{2.04}& 0.42& 0.63& \textbf{3.65}& 0.81\\
\bottomrule
 \end{tabular}
    \caption{Fidelity metrics of similarities on marginal distributions and joint distributions, analyzed in \ref{paragraph:fidelity-table-analysis}. Different models have different strengths, with the best-performed model being highlighted for each metric. CTGAN and TabAutoDiff show more balanced performance from the fidelity aspect (see section \ref{paragraph:fidelity-table-analysis}).}
    \label{tab:fidelity-table}
\end{table}

Among the models evaluated, TabAutoDiff and CTGAN emerge as standout performers. TabAutoDiff demonstrates a consistently balanced performance across all metrics, excelling in capturing numerical marginal distributions and category-to-category joint distributions. This balanced prowess suggests TabAutoDiff's ability to effectively replicate retail datasets' complex, inherent patterns. CTGAN, on the other hand, excels particularly in learning number-to-number and number-to-category interactions.  Both TabDDPM and SatAutoDiff models showed their strengths in learning categorical marginal distribution but failed to prove their intelligence in other distributions. We would not recommend AutoGAN from the fidelity perspective, because it did not stand out from any type of distribution examinations. \label{paragraph:fidelity-table-analysis}. GAN models can lead to poor representation of the relationships and correlations among columns when the generator fails to capture the full diversity and complexity of the original dataset because of Mode Collapse. This can happen especially when the category columns present imbalanced distributions. However, CTGAN employs techniques like mode-specific normalization to stabilize the learning process.

\begin{figure*}[!htb]
    \centering
    \includegraphics[width=1.0\textwidth]{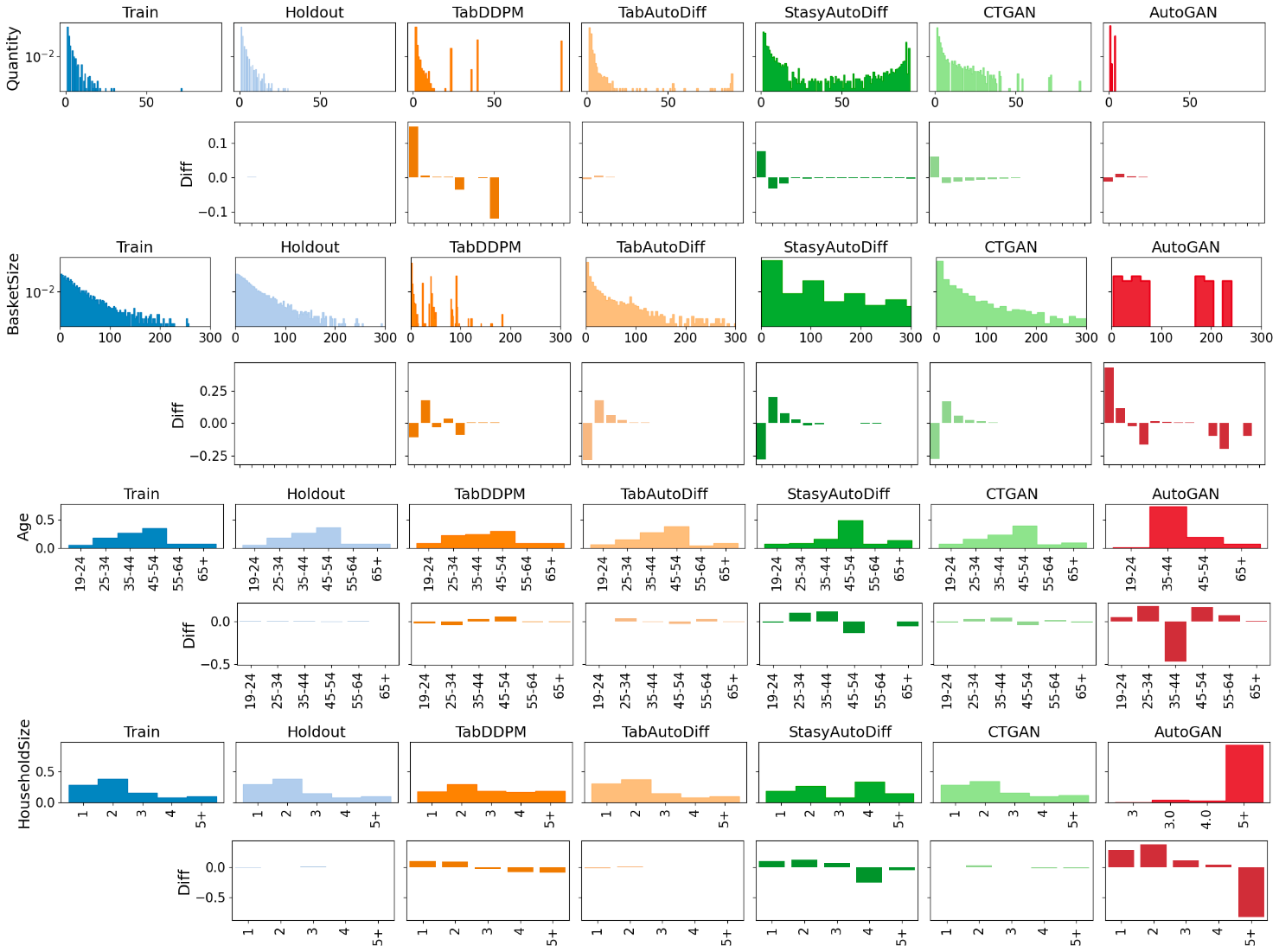}
    \caption{Distribution of feature columns from the training dataset, holdout dataset, and synthetic datasets, as well as the corresponding distribution difference to the one observed in the training dataset. The figure contains a primitive numerical column (Quantity), a derived numerical column (Basket Size), and primitive categorical columns (Age, Household Size), see \ref{paragraph:marginal-dist-analysis}.  Synthetic data generated by TabAutoDiff demonstrates feature distributions that closely mirror the ones of the original training dataset.}
    \label{fig:marginal-distribution}
\end{figure*}

\paragraph{\textbf{Marginal Distribution Visualization}} We brought back features from Figure \ref{fig:eda} and presented learned distributions from various generative models in  Figure \ref{fig:marginal-distribution}, plotting not only the feature distribution but also distributional differences between the real training data and the synthetic data. When it came to numerical features, quantity and basket size, TabAutoDiff and CTGAN were the only models that replicated the skewed distribution, though TabAutoDiff generated a small spike at the tail unexpectedly for the quantity distribution and CTGAN learned much fatter tails for both distributions. The diff plot proved These models' robustness again, where distributional difference histograms presented bars with height near 0. All models, except StasyAutoDiff and AutoGAN, were all capable to capture the right shape of category distributions, especially when the number of unique values in one category is low. However, if the dataset contains categorical columns with dramatic variation, we should expect a more concerning model performance. \label{paragraph:marginal-dist-analysis}
 
\begin{figure}[H]
    \centering
    \includegraphics[width=0.7\linewidth]{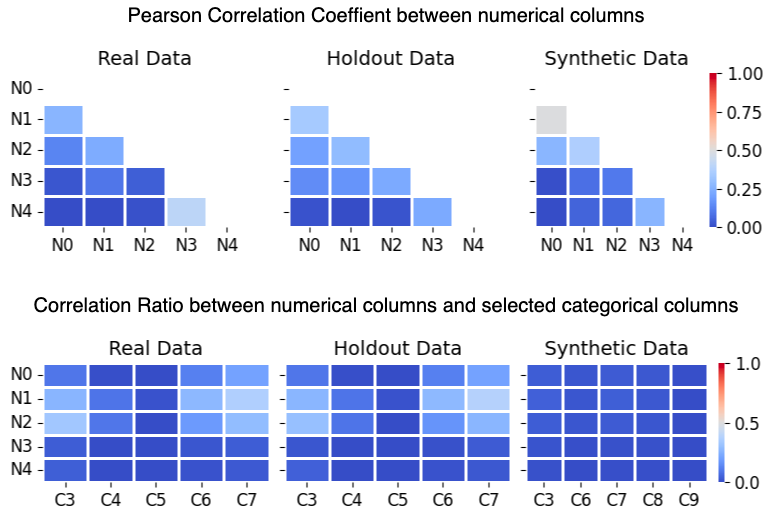}
    \caption{Heatmap of correlation metrics for Num-Num and Num-Cat interactions in the training, holdout, and CTGAN synthetic dataset. CTGAN model can replicate the feature interaction observed in the training dataset (see section \ref{paragraph:corr-matrix-analysis})}
    \label{fig:interaction_heatmap}
\end{figure}

\paragraph{\textbf{Correlation Matrix Visualization}} To build a more informative presentation, we specifically provided a more detailed presentation on CTGAN, which performed the best in replicating joint distributions. Though the real dataset shows a weak correlation between features, CTGAN exhibits remarkable strength in capturing both numerical interactions and mixed-type interactions in Figure \ref{fig:interaction_heatmap}, highlighting its capability to generate coherent and realistic relationships within the data. This makes CTGAN a top choice for applications where high fidelity in interaction data is critical. \label{paragraph:corr-matrix-analysis}

Though we recommended TabAutoDiff and CTGAN from the fidelity perspective, all generative models have metrics larger than the corresponding values from the holdout dataset. This indicates that there are still hidden patterns in the training dataset that these models are not fully replicating. Synthetic data generation for retail is inherently challenging due to the high degree of heterogeneity and the dynamic nature of customer shopping preferences. Due to the complexity and challenges in synthetic data generation, generative models still have further headroom to capture the latent structure of retail datasets fully. The proposed evaluation framework is pivotal in this regard, as it provides a standardized approach to assess model stability and performance. By enabling a consistent comparison across different models and metrics, this framework aids in understanding the nuances of each model's strengths and areas for improvement.

\subsection{Synthetic Data Utility Assessment}

\paragraph{\textbf{Classification Task}} To evaluate the model utility, we formulate two tasks. One is a classification task to identify premium customers who buy more products in one visit, predicting whether a customer will purchase more than 10 products based on their demographics and average unit price. We trained all classifiers supported by scikit-learn \cite{scikit-learn} and reported accuracy, F1, ROC, precision and recall of the model produces the highest accuracy, Bagging Classifier \cite{breiman1996bagging}, in Table \ref{tab:utility-table}. Among the synthetic data models evaluated, TabAutoDiff emerges as the best-performing model for the classification task, indicating TabAutoDiff's superior performance in generating useful synthetic data that can effectively train classification models. When comparing the utility metrics of the synthetic data generated by TabAutoDiff to those of the train and holdout datasets, it is evident that the model trained with synthetic data achieved similar performance on all metrics, which indicates the capability of synthetic data to generalize well to real data scenarios and serve as an effective proxy for real data. In this way, retailers can test marketing algorithms using synthetic data, eliminating the costs and risks of live A/B testing on real customers. \label{paragraph:classification-analysis}

\begin{table}[!htb]
    \centering\small
    \begin{tabular}[width=1.0\textwidth]{clllll}
\toprule
 & \multicolumn{5}{c}{Classification}\\
  \cmidrule(lr){2-6} 
 & Accuracy&   F1&ROC & Precision&Recall\\
\midrule
 Train& 0.65& 0.62 & 0.67 & 0.52 & 0.76\\
 Holdout& 0.66& 0.62 & 0.68 & 0.52 & 0.77 \\
 \midrule
 CTGAN& \textbf{0.68}& 0.40   & 0.60 & 0.63 & 0.29\\
 AutoGAN& 0.38& 0.53  &0.50 &0.37 &\textbf{0.96}\\
TabDDPM& 0.63&  0.11 & 0.51 &0.52 & 0.06\\
 StasyAutoDiff& 0.62& 0.16&  0.51& 0.45 & 0.10\\
 TabAutoDiff& \textbf{0.68}& \textbf{0.52} & \textbf{0.64}  & \textbf{0.59} & 0.47\\
 \bottomrule
 \end{tabular}
    \caption{Utility metrics on a classification task. TabAutoDiff model achieves the best performance from the utility aspect as it scores high on accuracy, F1, ROC, and precision. See detailed description in Sec.\ref{paragraph:classification-analysis}}
    \label{tab:utility-table}
\end{table}

\paragraph{\textbf{Product Association Analysis}} The other task is to analyze product association in the training, holdout and synthetic datasets. We performed market basket analysis at the product level to see if there is a significant affinity for products to be purchased together using the Apriori algorithm \cite{raschkas_2018_mlxtend}, a popular method used in data mining for extracting such association rules. Table \ref{tab:lift} presents the Lift metric, a measure of the likelihood that product B is bought when product A is bought, and the Conviction metric, which compares the probability that A appears without B if they were independent vs the actual frequency of A's appearance without B. 

\begin{align}
Confidence(A \rightarrow B) &= {P(A \cap B)}\Big/P(A) \\
Lift(A \rightarrow B) &= {Confidence(A \rightarrow B)}\Big/P(B) \\
Conviction(A\rightarrow B) &= (1 - P(B)) / (1 - Confidence(A \rightarrow B))
\end{align}

where $P(A\cap B)$ is probability both products being purchased, and$P(A), P(B)$ is the individual probability of purchasing product A or B accordingly. 

If Lift and Conviction much larger than 1, it means that product B is likely to be bought if product A is bought. In Table \ref{tab:lift}, both values for the synthetic datasets generated by AutoGAN and TabAutoDiff are significantly different from those of the Train and Holdout datasets. AutoGAN shows an exceptionally high lift and conviction values, indicating an overestimation of product pair occurrences, whereas TabAutoDiff’s lift, although lower, still does not align closely with the real datasets. The synthetic data generated by CTGAN, TabDDPM, and StasyAutoDiff did not observe any frequently purchased product pairs, indicating a fundamental gap in learning the essential co-occurrence relationships of products. This metric is invaluable for retailers as it helps identify product bundles, optimize placement strategies, and enhance cross-selling opportunities, thus leveraging consumer purchasing patterns to drive sales and customer satisfaction. By incorporating lift values, retailers can make data-driven decisions to refine their marketing and inventory strategies effectively. The evident difficulties reveal that existing generative models struggle in this aspect, which highlights the need for further refinement in replicating complex pairwise and higher-order relationships between products. \label{paragraph:product-association-analysis}
\begin{table}[h]
    \centering
    \begin{tabular}[width=1.0\linewidth]{cllll}
    \toprule
         &  Train &  Holdout& AutoGAN & TabAutoDiff\\
         \midrule
 Confidence& 0.18& 0.18& 0.99&\textbf{0.26}\\
    
         Lift& 1.73 & 1.72 & 20.89 & \textbf{4.83}\\
    
 Conviction& 1.10& 1.09& inf&\textbf{1.30}\\
    \bottomrule
    \end{tabular}
    \caption{Product association analysis for the listed synthetic data. Neither of the reported models can identify a similar product association rule observed in the training dataset. See analysis for each metric in Sec.\ref{paragraph:product-association-analysis}}
    \label{tab:lift}
\end{table}

\subsection{Synthetic Data Privacy Assessment}
Among the models we evaluated, TabAutoDiff showed a balanced performance in data privacy protection and model generalization. Table \ref{tab:privacy-table} presents privacy metrics for synthetic datasets generated by various models, focusing on the Distance to Closest Record (DCR) and the Closest Record Ratio (CCR). These metrics are crucial for assessing the privacy preservation capabilities of synthetic data, as they indicate how closely synthetic records resemble real training data points and the distribution balance, respectively.

Analyzing the DCR metric, the holdout set has the lowest DCR value, representing the benchmark distance within the real dataset. A larger DCR is preferred as it signifies that synthetic data points are not too close to any specific real training data points, thereby enhancing privacy. Among the models evaluated, TabAutoDiff demonstrates one of the higher DCR values. This indicates that its synthetic data maintains a significant privacy distance from the real data compared to other models, and stays at a lower risk of privacy leaking compared to the holdout dataset. TabAutoDiff also achieves the lowest CCR value, suggesting that the model did not overfit with the training data, thus providing good generalizability. \label{paragraph:privacy-analysis}

\begin{table}[H]
    \centering
    \begin{tabular}[width=1.0\linewidth]{ccc}
\toprule
 & DCR&   CCR\\
 \midrule
 Holdout & 1.0& - \\
 \midrule
 CTGAN& 4.24&  0.48\\
 AutoGAN& 8.82& 0.51\\
 TabDDPM& 4.52&  0.72\\
 StasyAutoDiff& 4.02& 0.54\\
 TabAutoDiff& \textbf{10.86}& \textbf{0.45}\\
 \bottomrule
\end{tabular}
    \caption{Privacy metrics of all tested models. TabAutoDiff stands out from the privacy aspect as it obtains the best performance in DCR and CCR. See detailed description in  Sec.\ref{paragraph:privacy-analysis}}
    \label{tab:privacy-table}
\end{table}

The retail industry is particularly cautious with customer data due to the sensitivity and privacy issues associated with handling such information. Strict regulations and the potential for reputational damage necessitate robust privacy preservation measures. The mixed performance of different models in terms of DCR and CCR highlights the need for a nuanced choice in synthetic data generation. While DCR is critical for ensuring individual data points are not too closely replicated, CCR helps ensure data generalizability, crucial for practical use in retail analytics. Thus, TabAutodiff is the best among evaluated models from the privacy perspective.

\section{Discussions}

Our comprehensive evaluation framework revealed distinct performances across various generative AI models in terms of fidelity, utility, and privacy for synthetic retail data. For fidelity, TabAutoDiff and CTGAN stood out, with TabAutoDiff demonstrating balanced performance across all metrics and CTGAN excelling in capturing joint distribution metrics, though all models showed room for improvement. In utility assessment, TabAutoDiff emerged as the top performer, effectively replicating the utility of real data in the classification task. However, none of the tested models demonstrate even a minimally acceptable performance in the product association analysis, indicating that further refinement of the model structure is necessary to achieve satisfactory results. For privacy, Distance to Closest Record (DCR) and Closest Cluster Ratio (CCR) metrics highlighted the models' ability to anonymize data effectively and balance generalization and identified TabAutoDiff again as the best one among the tested models. The proposed evaluation framework successfully highlighted the strengths and areas for improvement of each model, promoting the development of more effective and reliable synthetic data generation techniques for the retail sector.

Overall, our evaluation framework offers a standardized approach to assessing synthetic data generation models, facilitating consistent and transparent benchmarking. This can drive further research and development in the field, encouraging the creation of more sophisticated models capable of better capturing the complexities of retail data. To improve the evaluation framework, future research could focus on developing domain-specific metrics that capture the unique characteristics and complexities of various retail datasets, such as transactional data, inventory data, etc. Expanding the diversity and size of the datasets used for evaluation, possibly through collaboration with industry partners or data-sharing initiatives, would enhance the robustness and generalizability of the evaluation framework's findings. 

Anticipated advancements in synthetic data generation and evaluation include the development of more sophisticated generative models or LLM models capable of capturing higher-order dependencies and dynamic patterns presented in retail data. As these models evolve, they will not only improve in faithfully replicating the complexities of consumer behaviors but also in their ability to fill gaps where real-world data might be sparse or biased. Once the model reaches a mature level of learning all the patterns that retailers care about, we can confidently say that the generative model mirrors real customer purchase behavior. This validation would open up a multitude of applications, allowing the model to be used for inference such as demand forecasting and dynamic pricing. Furthermore, the robust nature of such advanced models could be leveraged within simulation environments to develop simulated A/B testing and test variations of a product or service in a controlled and cost-effective manner. A simulation environment with reliable generative models can also bridge the gap of reinforcement learning (RL) agents deployment aimed at personalized coupon-targeting strategies and optimizing customer engagement \cite{xia2024simulationbased, ie_recsim_2019}. By aligning synthetic data generation advancements with these application areas, businesses can not only gain deeper insights but also implement more dynamic and responsive retail strategies

\clearpage

%\clearpage
\baselineskip=13pt
\bibliographystyle{plain}
\nocite{*}
\bibliography{ref}

%\clearpage
\appendix

\section{Column name mapping}\label{appendix:column-mapping}
The complete journey data has relative long names for each column. For a succinct presentation of the correlation heatmap in figure \ref{fig:corr-eda}, we created a column mapping to label numerical columns and categorical columns.

\begin{table}[H]
    \centering
    \begin{tabular}{lcc}
    \toprule
          Raw name&Type&  Label\\
          \midrule
          product\_id & categorical & C1\\
          household\_id & categorical & C2\\
          week & categorical & C3\\
          manufacturer\_id & categorical & C4\\
          department & categorical & C5\\
          brand & categorical & C6\\
          product\_category & categorical & C7\\
          product\_type & categorical & C8\\
          package\_size & categorical & C9\\
          age & categorical & C10\\
          homeownership & categorical & C11\\
          marital\_status & categorical & C12\\
          household\_size & categorical & C13\\
          household\_comp & categorical & C14\\
          kids\_count & categorical & C15\\
          \midrule
          quantity & numerical & N1\\
          sales\_value & numerical & N2\\
          retail\_disc & numerical & N3\\
          coupon\_disc & numerical & N4\\
          coupon\_match\_disc & numerical & N5\\
          unit\_price & numerical & N6\\
          \bottomrule
    \end{tabular}
    \caption{Column name mapping to the shortened label.}
    \label{tab:column-mapping}
\end{table}

\end{document}

%% file: def-smile.tex
\numberwithin{example}{section}

%% file: front.tex
\usepackage[utf8x]{inputenc}
\usepackage{mathrsfs}
\usepackage{amssymb}
\usepackage{amsfonts}
\usepackage{amsmath}
\usepackage{amsthm,verbatim}
\usepackage{wrapfig}
\usepackage{multirow}
\usepackage{longtable}
\usepackage{enumerate}
\usepackage{color, graphicx}
\usepackage{tikz}

\def\boxit#1{\vbox{\hrule\hbox{\vrule\kern6pt
          \vbox{\kern6pt#1\kern6pt}\kern6pt\vrule}\hrule}}